\pdfoutput=1
\documentclass[11pt]{article}

\usepackage{EMNLP2023}

\usepackage{times}
\usepackage{latexsym}

\usepackage[T1]{fontenc}
\usepackage[utf8]{inputenc}

\usepackage{microtype}

\usepackage{inconsolata}

\usepackage[colorinlistoftodos,prependcaption,textsize=tiny]{todonotes}
\usepackage{tablefootnote}
\usepackage{bm}
\usepackage{dsfont}
\usepackage{amsmath,amssymb}
\usepackage{mathtools}
\usepackage{booktabs}
\usepackage{multirow}
\usepackage{graphicx} 
\usepackage{float} 
\usepackage{subfigure} 
\usepackage{enumitem}
\setlist[description]{format=\textcolor{blue},parsep=2pt, itemsep=1pt}
\newcommand{\ST}{{\text{ST}}}
\newcommand{\ED}{{\text{ED}}}
\def \FFN{\text{FFN}}

\def \sig{\text{Sigmoid}}
\def \softmax{\text{Softmax}}

\newcommand{\maxpool}{\text{Maxpooling}}

\def \min{\text{min}}
\def \max{\text{max}}

\newcommand{\scierc}{SciERC}
\newcommand{\ace}{ACE2005}
\newcommand{\acef}{ACE2004}
\def \scibert{\textit{scibert-scivocab-uncased}}
\def \bertbase{\textit{bert-base-uncased}}
\def \albert{\textit{albert-xxlarge-v1}}

\newcommand{\bertb}{$\text{BERT}_B$}

\newcommand{\scib}{SciBERT}
\newcommand{\alb}{ALBERT}

\newcommand{\bkb}{\texttt{Backbone}}
\newcommand{\gcn}{\texttt{GCN}}
\newcommand{\mfvi}{\texttt{MFVI}}
\newcommand{\hgnn}{\texttt{HGERE}}

\def \Ter{\textit{ternary}}
\def \Sib{\textit{sibling}}
\def \Cop{\textit{co-parent}}
\def \Gp{\textit{grand-parent}}
\def \ter{\textit{ter}}
\def \sib{\textit{sib}}
\def \cop{\textit{cop}}
\def \gp{\textit{gp}}
\def \tersib{\textit{tersib}}
\def \tercop{\textit{tercop}}
\def \tergp{\textit{tergp}}
\def \tersibcop{\textit{tersibcop}}
\def \tersibcopgp{\textit{tersibcopgp}}
\def \tersibgp{\textit{tersibgp}}
\def \tercopgp{\textit{tercopgp}}
\def \sors{\text{sub-obj-rel}}
\def \rrs{\text{rel-rel}}

\def \ztt{\texttt{z}}
\def \nertype{\mathcal{C}_e}
\def \retype{\mathcal{C}_r}
\def \nul{\texttt{null}}

\def \ent{\text{Ent}}
\def \rel{\text{Rel}}
\def \relp{\text{Rel+}}

\def\gE{{\mathcal{E}}}

\def\gG{{\mathcal{G}}}

\def\gN{{\mathcal{N}}}

\def\gV{{\mathcal{V}}}

\def\rvf{{\mathbf{f}}}
\def\rvg{{\mathbf{g}}}
\def\rvh{{\mathbf{h}}}
\def\rvu{{\mathbf{i}}}

\def\rvm{{\mathbf{m}}}

\def\rvu{{\mathbf{u}}}

\def\rvw{{\mathbf{w}}}
\def\rvx{{\mathbf{x}}}

\def\rvxs{{\mathbf{xs}}}
\def\rvxe{{\mathbf{xe}}}

\def\rvW{{\mathbf{W}}}

\def\rvhr{{\mathbf{hr}}}
\def\rvhs{{\mathbf{hs}}}
\def\rvho{{\mathbf{ho}}}

\def\emW{{W}}

\newcommand{\biaf}{\text{biaf}}
\newcommand{\attn}{\text{attn}}

\newcommand{\bss}{\backslash S}
\newcommand{\bso}{\backslash O}

\newcommand{\lessdis}{\vspace{-0.3cm}}

\title{Joint Entity and Relation Extraction with Span Pruning and \\  Hypergraph Neural Networks}

\author{$\text{Zhaohui Yan}^{1,2}$, $\text{Songlin Yang}^3$\thanks{\; This work was done when Songlin was at ShanghaiTech.}, $\text{Wei Liu}^{1,2}$, $\text{Kewei Tu}^{1,2}$\thanks{\; Corresponding Author} \\
  $^{1}\text{School of Information Science and Technology, ShanghaiTech University}$ \\
    $^{2}\text{Shanghai Engineering Research Center of Intelligent Vision and Imaging}$\\ 
    $^{3}\text{MIT CSAIL}$\\ 
    {\tt \{yanzhh, liuwei4, tukw\}@shanghaitech.edu.cn}\\{\tt yangsl66@mit.edu}\\
  }
\begin{document}
\maketitle
\begin{abstract}
Entity and Relation Extraction (ERE) is an important task in information extraction. Recent marker-based pipeline models achieve state-of-the-art performance, but still suffer from the error propagation issue. Also, most of current ERE models do not take into account higher-order interactions between multiple entities and relations, while higher-order modeling could be beneficial.In this work, we propose HyperGraph neural network for ERE ($\hgnn{}$), which is built upon the PL-marker (a state-of-the-art marker-based pipleline model). To alleviate error propagation,we use a high-recall pruner mechanism to transfer the burden of entity identification and labeling from the NER module to the joint module of our model. For higher-order modeling, we build a hypergraph, where nodes are entities (provided by the span pruner) and relations thereof, and hyperedges encode interactions between two different relations or between a relation and its associated subject and object entities. We then run a hypergraph neural network for higher-order inference by applying message passing over the built hypergraph. Experiments on three widely used benchmarks (\acef{}, \ace{} and \scierc{}) for ERE task show significant improvements over the previous state-of-the-art PL-marker. \footnote{Source code is availabel at \url{https://github.com/yanzhh/HGERE}}
\end{abstract}

%===============================================================
\section{Introduction}

Entity and Relation Extraction (ERE) is a fundamental task in information extraction (IE), compromising two sub-tasks: Named Entity Recognition (NER) and Relation Extraction (RE).  There is a long debate on joint vs. pipeline methods for ERE. Pipeline decoding extracts entities first and  predicts relations solely on pairs of extracted entities, while joint decoding predicts entities and relations simultaneously.

Recently, the seminal work of \cite{zhong-chen-2021-frustratingly} shows that pipeline decoding with a frustratingly simple marker-based encoding strategy --- i.e., inserting solid markers \cite{baldini-soares-etal-2019-matching, xiao-etal-2020-denoising} around predicted subject and object spans in the input text --- achieves state-of-the-art RE performance.  Modified sentences (with markers) are fed into powerful pre-trained  large language models (LLM) to obtain more subject- and object-aware representations for RE classification, which is the key to the performance improvement. However, current marker-based pipeline models (e.g., the recent state-of-the-art ERE model PL-marker \cite{ye-etal-2022-packed}) \emph{only} send predicted entities from the NER module to the RE module, therefore missing entities would never have the chance to be re-predicted, suffering from the \emph{error propagation issue}.
On the other hand, for joint decoding approaches (e.g. Table Filling methods \cite{miwa-sasaki-2014-modeling, zhang-etal-2017-end, wang-lu-2020-two})---though they do not suffer from the error propagation issue---it is hard to incorporate markers for leveraging LLMs, since entities are not predicted prior to relations. Our desire is to obtain the best of two worlds, being able to use marker-based encoding mechanism for enhancing RE performance and meanwhile alleviating the error propagation problem. We adopt PL-marker as the backbone of our proposed model and a span pruning strategy to mitigate error propagation. That is, instead of sending only predicted entity spans to the RE module, we \emph{over-predict} candidate spans so that the recall of gold entity spans is nearly perfect (but there also could be many non-entity spans),  transferring the burden of entity classification and labeling from the NER module to the RE module of PL-marker. The number of over-predicted spans is upper-bounded, balancing the computational complexity of marker-based encoding and the recall of gold entity span. Empirically, we find this simple strategy by itself clearly improves PL-marker.

We further incorporate a \emph{higher-order} interaction module into our model. Most previous ERE models either implicitly model the interactions between instances by shared parameters \cite{wang-lu-2020-two,yan-etal-2021-partition,wang-etal-2021-unire} or 
use a traditional graph neural network that models pairwise connections between a relation and an entity \cite{sun-etal-2019-joint}. 
It is difficult for these approaches to explicitly model higher-order relationships among multi-instances, e.g. the dependency among a relation and its corresponding subject and object entities. Many recent works in structured prediction tasks show that explicit higher-order modeling is still beneficial even with powerful large pretrained encoders \cite[][\textit{inter alia}]{zhang-etal-2020-efficient, li-etal-2020-high, yang-tu-2022-combining, zhou-etal-2022-fast}, motivating us to use an additional higher-order module to enhance performance.

 A common higher-order modeling approach is by means of probabilistic modeling (i.e., conditional random field (CRF)) with end-to-end Mean-Field Variational Inference (MFVI), which can be seamlessly integrated into neural networks as a recurrent neural network layer \cite{DBLP:conf/iccv/0001JRVSDHT15}, and has been widely used in various structured prediction tasks, such as dependency parsing \cite{wang-etal-2019-second}, semantic role labeling \cite{li-etal-2020-high,zhou-etal-2022-fast}, and information extraction \cite{Jia2022JointIE}. However, the limitations of CRF modeling with MFVI are i): CRF's potential functions are parameterized in log-linear forms with strong independence assumptions, suffering from low model capacities \cite{qu2022neural},  ii) MFVI uses fully-factorized Bernoulli distributions to approximate the otherwise multimodal true posterior distributions, oversimplifying the inference problem and thus is sub-optimal. Therefore we need more \emph{expressive} tools to improve the quality of higher-order inference. Fortunately, there are many recent works in the machine learning community showing that graph neural networks (GNN) can be used as an inference tool and outperform approximate statistical inference algorithms (e.g., MFVI) \cite{DBLP:conf/iclr/YoonLXZFUZP18,Zhang2020FactorGN,DBLP:conf/nips/KuckCTLSSE20,DBLP:conf/aistats/SatorrasW21} (see \cite{Hua2022GraphNN} for a survey). Inspired by these works, we employ a hypergraph neural network (HyperGNN) instead of MFVI for high-order inference and propose our model \hgnn{} (HyperGraph Neural Network for ERE).
Concretely, we build a hypergraph where nodes are candidate subjects and objects (obtained from the span pruner) and relations thereof, and hyperedges encode the interactions between either two relations with shared entities or a relation and its associated subject and object entity spans. In contrast, existing GNN models for IE \cite{sun-etal-2019-joint,nguyen-etal-2021-cross} only model the pairwise interactions between a relation and one of its corrsponding entity. We empirically show the advantages of our higher-order interaction module (i.e., hypergraph neural network) over MFVI and tranditional GNN models. 

Our contribution is three-fold: i) We adopt a simple and effective span pruning method to mitigate the error propagation issue, enforcing the power of marker-based encoding.  ii) We propose a novel hypergraph neural network enhanced higher-order model, outperforming higher-order CRF-based models with MFVI. iii) We show great improvements over the prior state-of-the-art PL-marker on three commonly used benchmarks for ERE: \acef{}, \ace{} and \scierc{}.

%===============================================================
\section{Background}

\begin{figure*}[!ht]
% \lessdis{}
    \centering
    \includegraphics[width=0.9\textwidth]{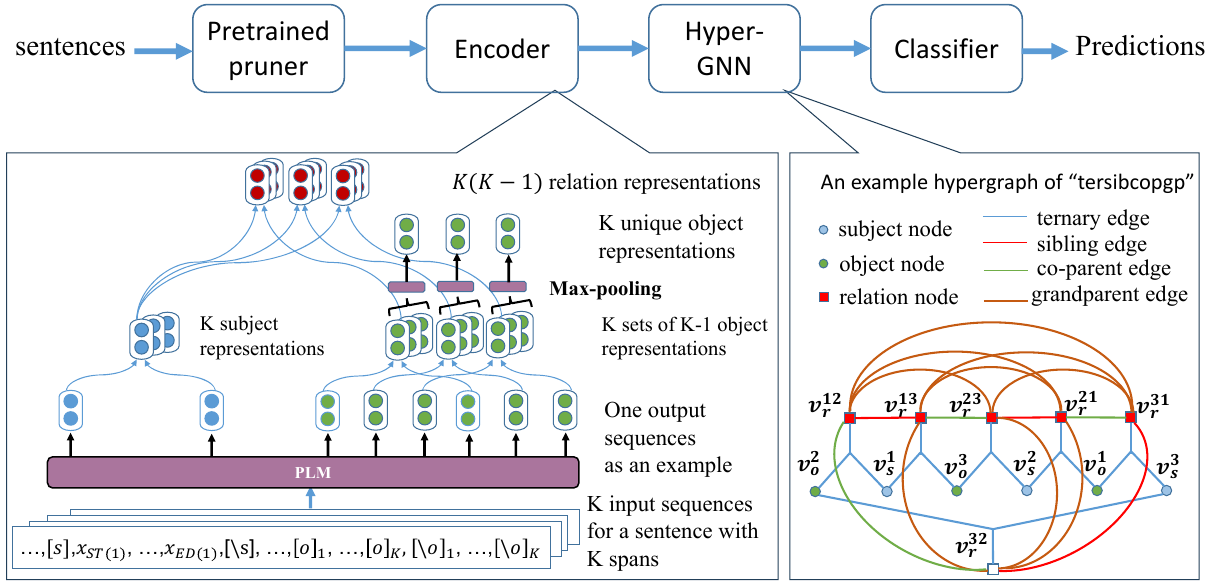} 
    \caption{Illustration of our framework}
    \label{fig.model} 
    % \lessdis{}
\end{figure*}

\subsection{Problem formulation}
Given a sentence $X$ with $n$ tokens: $x_1, x_2, ..., x_n$,  an entity span is a sequence of tokens labeled with an entity type and a relation is an entity span pair labeled with a relation type. We denote the set of all entity spans of the sentence with a span length limit $L$ by $S(X)=\{s_1, s_2,...,s_m\}$ and define $\ST(i)$ and $\ED(i)$ as the start and end token indices of the span $s_i$.

The joint ERE task is to simultaneously solve the NER and RE tasks. Let $\nertype{}$ be the set of entity types and $\retype{}$ be the set of relation types. For each span $s_i\in S(X)$, the NER task is to predict an entity type $y_e(s_i)\in \nertype{}$ or $y_e(s_i)=\nul{}$ if the span $s_i$ is not an entity. The RE task is to predict a relation type $y_r(r_{ij})\in \retype{}$ or $y_r(r_{ij})=\nul{}$ for each span $r_{ij}=(s_i, s_j), s_i, s_j\in S(X)$.

\subsection{Packed levitated marker (PL-marker)}
\citet{zhong-chen-2021-frustratingly} insert two pairs of solid markers (i.e., $[S]$ and $[\bss]$) to highlight both the subject and object entity spans in a given sentence, and this simple approach achieves state-of-the-art RE performance. We posit that this is because LLM is more aware of the subject and object spans (with markers) and thus can produce better span representations to improve RE. But this strategy needs to iterate over all possible entity span pairs and is thus slow. To tackle the efficiency problem, \citet{zhong-chen-2021-frustratingly} propose an approximated variant wherein each possible entity span is associated with a pair of levitated markers (i.e., $[O]$ and $[\bso]$) whose representations are initialized with the positional embedding of the span's start and end tokens, and all such levitated markers are concatenated to the end of the sentence. As such, levitated-marker-based encoding needs only one pass, significantly improving efficiency at the cost of slight performance drop. \citet{zhong-chen-2021-frustratingly} also propose a masked attention mechanism such that the original input text tokens are not able to attend to markers, while markers can attend to paired markers (but not unpaired markers) and all input text tokens. As a consequence, the relative positions of levitated markers in the concatenated sentence do not matter at all, eliminating potential implausible inductive bias on the  concatenation order.

However, marker-base encoding is only used in RE, not in NER. To leverage marker-based encoding in the NER module for modeling span interrelations, PL-marker \cite{ye-etal-2022-packed} associates each possible span with two levitated markers and concatenates all of them to the end of the input sentence. However, this strategy could make the input sentence extremely long since there are quadratic number of spans. To solve this issue, PL-marker clusters the markers based on the starting position of their corresponding spans, and divides them into $N$ groups. Then the input sentence is duplicated $N$ times and each group of levitated markers is concatenated to the end of one sentence copy. \citet{ye-etal-2022-packed} refers to this strategy as \emph{neighborhood-oriented packing scheme}. Furthermore, to balance the efficiency and the model expressiveness, \citet{ye-etal-2022-packed} combine solid markers and levitated markers, proposing \emph{Subject-oriented Packing for Span Pair} in the RE module. That is, if there are $m$ entities, they copy the sentence for $m$ times, and for each copy, they use solid markers to mark a different entity as the subject and concatenate the levitated markers of all other entities (as objects) at the end of the sentence. 

%===============================================================

\section{Method}
\paragraph{Overview.} Our method is built upon the state-of-the-art PL-marker. 
% We use a span pruner to over-predict candidate entity spans similar to the NER module of PL-marker. But it aims to prune unlikely entities instead of predicting likely entities to achieve higher recall.
We employ a high-recall span pruner to obtain candidate entity spans, similar to the NER module in PL-marker. However, instead of aiming to accurately predict all possible entity spans, our pruner focuses on removing unlikely candidates to achieve a much higher recall.
Then we feed the candidate span set to the RE module to obtain entity and relation representations, which are used to initialize the node representations of our hypergraph neural network for higher-order inference with a message passing scheme. Finally, we perform NER and RE based on the refined entity and relation representations. Fig.~\ref{fig.model} depicts the neural architecture of our model.

%------------------------------------
\subsection{Span Pruner}
We adopt the neighborhood-oriented packing scheme from PL-marker for span encoding, except that we simply predict entity existence (i.e., binary classification) instead of predicting entity labels during the training phrase. See Appendix~\ref{appd:pruner} for details. 

To produce a candidate span set, we rank all the spans by their scores and take top $K$ as our prediction $S_p(X)$. We assume that the number of entity spans of a sentence is linear to its length $n$, so $K$ is set to $\lambda \cdot n$ where $\lambda$ is a coefficient. For a very long sentence, the number of entity spans is often sublinear to $n$, while for a very short sentence, we wish to keep enough candidate spans, so we additionally set an upper and lower bound: $K = \max(l_\min, \min(\lambda \cdot n, l_\max))$. 

In practice, with our span pruner, more than 99\% gold entity spans are included in the candidate set for all three datasets. If we predict entities as in PL-marker instead of pruning, only around 95\% and 80\% gold entities are kept in the predicted entities for \ace{} and \scierc{} respectively, leading to severe error propagation (see \S{}\ref{sec.eff_pruner} for an ablation study).

The span pruner is trained independently from the joint ERE model  introduced in the next section. This is because the joint ERE training loss will be defined based on candidate entity spans produced by the span pruner. When sharing parameters, the pruner would provide a different candidate span set during training, leading to moving targets and thereby destabilizing the whole training process.

%------------------------------------
\subsection{Joint ERE Model: First-order Backbone }
\label{sec:backbone}
The backbone module is based on the RE module of PL-marker. Concretely, given an input sentence $X=\{x_1, x_2,..., x_n\}$ and a subject span $s_i=(x_{\ST(i)}, x_{\ED(i)})\in S_p(X)$ provided by the span pruner, every entity span $s_j\in S_p(X), 1\le j\le K, j\neq i$ could be a candidate object span of $s_i$. 
The module inserts a pair of solid markers $[S]$ and $[\bss]$ before and after the subject span and assign every object span $s_j$ a pair of levitated markers $[O]_j$ and $[\bso]_j$. As shown below, the levitated markers are packed together and inserted at the end of the input sequence to a PLM:
\vspace{-0.1cm}
\begin{align*}
 x_1, ..., [S], x_{\ST(i)}, ..., x_{\ED(i)}, [\bss], ..., x_n, \\
 [O]_1, ..., [O]_K, [\bso]_1, ..., [\bso]_K
\end{align*}
Then we obtain the contextualized hidden representation $\rvh_x$ of the modified input sequence and the final subject representation is:
\vspace{-0.1cm}
$$
\rvh_s(s_i)=\FFN_s([\rvh_x([S]);\rvh_x([\bss])])
$$
FFN represents a single linear layer in this work. The object representation of $s_j$ for the current subject $s_i$ and the representation of relation $r_{ij}=(s_i, s_j)$ are:
\vspace{-0.1cm}
$$
\rvh_o^i(s_j)=\FFN_o([\rvh_x([O]_j);\rvh_x([\bso]_j)])
$$
$$
\rvh_r(r_{ij})=\FFN_r([\rvh_s(s_i);\rvh_o^i(s_j)])
$$
Repeating $K$ times, we get all $K$ subject representations and $K(K-1)$ relation representations. As the object representation of $s_j$ is not identical for different subject span $s_i$, there are $K$ object representation sets $\rvh_o^i, 1\le i\le K$.
We apply a max-pooling layer to obtain a unique object representation for each object span $s_j\in S_p(X)$:
$$
\rvh_o(s_j) = \maxpool_{1\le i\le K, i\neq j}(\rvh_o^i(s_j))
$$

%-----------------------------------------
\subsection{Joint ERE Model: Higher-order Inference with Hypergraph Neural Networks}

\paragraph{Hypergraph Building}
So far, the representations of the entities and relations from the backbone module do not explicitly consider beneficial interactions among related instances. To model higher-order interactions among a relation and its associated subject and object entities as well as between any two relations sharing an entity, we build a hypergraph $\gG=(\gV,\gE)$ to connect the related instances.The nodes set $\gV$ is composed of candidate subjects, objects (provided by the span pruner) and all possible pairwise relations thereof, and we denote them as $\gV_s=\{v_s^i|i\in[1,K]\}$, $\gV_o=\{v_s^j|j\in[1,K]\}$ and $\gV_r=\{v_r^{ij}|i,j\in[1,K],i\neq j\}$. 

Hyperedges $\gE$ capture the interactions we are concerned with, and they can be divided into two categories: the subject-object-relation (\sors{}) hyperedges $\gE_{sor}$ and the relation-relation (\rrs{}) hyperedges $\gE_{rr}$. Each hyperedge $e^{ij}_{sor}\in \gE_{sor}$ connects a subject node $v_s^i$, an object node $v_o^j$ and the corresponding relation node $v_r^{ij}$, and we refer to these hyperedges as \Ter{} edges (\ter{} for short). Each \rrs{} edge $e^{ijk}_{rr}\in \gE_{rr}$ connects two relation nodes with a shared subject or object entity. We assume in a relation, the subject is the parent node and the object is the child node, and then we can refine \rrs{} edges into three subtypes, $\Sib{}$ ($\sib{}$, connecting $v_r^{ij}$ and $v_r^{ik}$) , $\Cop{}$ (\cop{}, connecting $v_r^{ij}$ and $v_r^{kj}$) and $\Gp{}$ ($\gp{}$, connecting $v_r^{ij}$ and $v_r^{jk}$), following the common definitions in the dependency parsing literature. 

If we incorporate all aforementioned hyperedges into the hypergraph, we obtain the $\tersibcopgp{}$ variant which is illustrated in Fig.~\ref{fig.model}. By removing some types of hyperedges we can get different variants, but without loss of generality we describe the message passing scheme in the following using $\tersibcopgp{}$.  

% As such, we can define a conditional random field (CRF) on the hypergraph  by assigning potential scores for each hyperedge and leverage probabilistic inference algorithms such as MFVI for higher-order inference. However, as discussed in \S{}1, we want to use a more expressive method to improve  inference quality and introduce a HyperGraph Neural Network (HGNN) as described next. 

As such, we can define a CRF on the hypergraph and leverage probabilistic inference algorithms such as MFVI for higher-order inference. However, as discussed in \S{}1, we can use a more expressive method to improve  inference quality and introduce a HyperGraph Neural Network (HGNN) as described next.

%---------------------------------------------

\paragraph{Initial node representation}
For a relation node $v_r^{ij}$ with its associated subject node $v_s^i$ and object node $v_o^j$, we use $\rvg^{l}(v_r^{ij}), \rvg^{l}(v_s^i)$, $\rvg^{l}(v_o^j)$ to denote their respective representation outputs from the $l$-th HGNN layer.  Initial node representations (before being fed to a HGNN) are $\rvg^{0}(v_s^i)=\rvh_s(s_i)$, $\rvg^{0}(v_o^j)=\rvh_o(s_j)$ and $\rvg^{0}(v_r^{ij})=\rvh_r(r_{ij})$, respectively (from the backbone module).

\paragraph{Message representation}
A hyperedge connecting to nodes serve as the bridge for message passing between nodes connected by it. Let $\mathcal{N}_{e}(v)$ be the set of hyperedges connecting to a node $v$.

For a $\ter{}$ hyperedge $e_{ter}^{ij}\in \gE_{sor}$ connecting  a subject node $v_s^i$, a object node $v_o^j$ and a relation node $v_r^{ij}$, the message representation it carries is:
\vspace{-0.1cm}
\begin{align*}
    \rvhr^{l}_{ij} & = \FFN^{ter}_r(\rvg^{l-1}(v_r^{ij})) \\
     \rvhs^{l}_{i} & = \FFN^{ter}_s(\rvg^{l-1}(v_s^i)) \\
     \rvho^{l}_{j} & = \FFN^{ter}_o(\rvg^{l-1}(v_o^j)) \\
     \rvm^{l}(e_{ter}^{ij}) & =\FFN^{ter}_e( \rvhr^{l}_{ij} \circ \rvhs^{l}_{i}  \circ\rvho^{l}_{j})
\end{align*}
where $\circ$ is the Hadamard product.

A \rrs{} edge $e_{\ztt}^{ijk}\in \gE_{rr}, \ztt{}\in \{\sib, \cop, \gp{}\}$ connects two relations sharing an entity. For simplicity, we denote them relation $a$ and $b$. If we fix $a$ as $a \triangleq v_r^{ij}$, then as 
previously described, relation $b$ is $v_r^{ik}$ for $\sib{}$ edge, $v_r^{kj}$ for $\cop{}$ edge, and  $v_r^{jk}$ for $\gp{}$ edge.
The message $e_{\ztt}^{ijk}$ carries is given by,
\vspace{-0.1cm}
\begin{align*}
& \rvh^{l}_{\ztt}(a)  = \FFN^{\ztt}_a(\rvg^{l-1}(a)) \\
& \rvh^{l}_{\ztt}(b) = \FFN^{\ztt}_b(\rvg^{l-1}(b)) \\
& \rvm^{l}(e_{\ztt}^{ijk})  =\FFN^{\ztt}_e( \rvh^{l}_{\ztt}(a) \circ \rvh^{l}_{\ztt}(b))
\end{align*}

\paragraph{Node representation update}
We aggregate messages for each node $v\in \gV$  from adjacent edges $\mathcal{N}_e(v)$  with an attention mechanism by taking a learned weighted sum, and add the aggregated message to the prior node representation, 
\vspace{-0.1cm}
\begin{align*}
   & \beta^l(e, v) = \rvw^{\top}\sigma(\rvW[\rvg^{l-1}(v); \rvm^{l}(e)]  \\
   & \alpha^l(e, v) = \frac{\exp{\beta^l(e, v)}}{\sum_{e'\in \mathcal{N}_e(v)}\exp{\beta^l(e', v)}} \\
   & \rvg^{l}(v) =  \rvg^{l-1}(v) + \sum_{e\in N_e(v)} \alpha^l(e, v)\rvm^{l}(e)
\end{align*}
where $\sigma(\cdot)$ is a non-linear activator and $\rvw, \rvW$ are two trainable parameters. An entity node would receive messages only from $\ter{}$ edges  while a relation node would receive messages from both $\ter{}$ edges and rel-rel edges.

\paragraph{Training}
We obtain refined $\rvg^l(v)$ from the final layer of HGNN. Give an entity span $s_i\in S_p(X)$, we concatenate the corresponding subject representation $\rvg^l(v_s^i)$ and object representation $\rvg^l(v_o^i)$ to obtain the entity representation, and compute the probability distribution over the types $\{\nertype\}\bigcup \{\nul{}\}$:
\vspace{-0.1cm}
$$
P_e(\hat{y}_e|s_i)=\softmax(\FFN_e^{cls}([\rvg^{l}(v_s^i);\rvg^{l}(v_o^i)]))
$$
Given a relation $r_{ij}=(s_i,s_j), s_i,s_j\in S_p(X)$, we compute the probability distribution over the types $\{\retype\}\bigcup \{\nul{}\}$:
$$
P_r(\hat{y}_r|r_{ij}) = \softmax(\FFN_r^{cls}(\rvg^{l}(v_r^{ij})))
$$
We use the cross-entropy loss for both entity and relation prediction:
\vspace{-0.1cm}
\begin{align*}
    L_e & = -\sum_{s_i\in S_p(X)} \log(P_e(y_e^*(s_i)|s_i)) \\
    L_r & = -\sum_{s_i,s_j\in S_p(X)} \log(P_r((y_r^*(r_{ij})|r_{ij}))
\end{align*}
where $y_e^*$ and $y_r^*$ are gold entity and relation types respectively. The total loss is $L=L_e + L_r$.

%=============================================================

\section{Experiment}

%-------------------- main table------------------------------------
\begin{table*}[!ht]
% \linespread{0.95}
% \lessdis{}
\small
\centering
\begin{tabular}{lcccccccccc} 
\toprule
\multirow{2}{*}{Models}                   & \multirow{2}{*}{Encoder}                                                      & \multicolumn{3}{c}{\ace{}}                   & \multicolumn{3}{c}{\acef{}}                   & \multicolumn{3}{c}{\scierc{}}                     \\ 
\cmidrule{3-11}
                                          &                                                                               & Ent           & Rel           & Rel+          & Ent           & Rel           & Rel+          & Ent           & Rel           & Rel+           \\ 
\midrule
% \cite{li-etal-2019-entity}                      & \multirow{2}{*}{Bert large}                                                   & 84.8          & -             & 60.2          & 83.6          & -             & 49.4          & -             & -             & -              \\
% \cite{lin-etal-2020-joint}                       &                                                                               & 88.8          & 67.5          & -             & -             & -             & -             & -             & -             & -              \\ 
% \midrule
\cite{wadden-etal-2019-entity}$^{\star}$                & \multirow{11}{*}{\begin{tabular}[c]{@{}c@{}}\bertb{}/\\ \scib{}\end{tabular}} & 88.6          & 63.4          & -             & -             & -             & -             & 67.5          & 48.4          & -              \\
\cite{wang-etal-2021-unire}$^{\star}$                    &                                                                               & 88.8          & -             & 64.3          & 87.7          & -             & 60.0          & 68.4          & -             & 36.9           \\
\cite{zhong-chen-2021-frustratingly}$^{\star}$           &                                                                               & 90.1          & 67.7          & 64.8          & 89.2          & 63.9          & 60.1          & 68.9          & 50.1          & 36.8           \\
\cite{yan-etal-2021-partition}                    &                                                                               & -             & -             & -             & -             & -             & -             & 66.8          & -             & 38.4           \\
\cite{shen2021trigger}$^{\star}$                        &                                                                               & 87.6          & 66.5          & 62.8          & -             & -             & -             & 70.2          & 52.4          & -              \\
\cite{nguyen-etal-2021-cross}                    &                                                                               & 88.9          & 68.9          & -             & -             & -             & -             & -             & -             & -              \\
\cite{ye-etal-2022-packed}$_{\textit{re}}^{\star}$    &                                                                               & 89.1          & 68.3          & 65.1          & 88.5          & 66.3          & 62.2          & 68.8          & 51.1          & 38.3           \\
\bkb{}$^{\star}$                                &                                                                               & 90.0          & 69.8          & 66.7          & 89.5          & 66.6          & 62.1          & 71.3          & 52.3          & 40.2           \\
\gcn{}$^{\star}$                                     &                                                                               & 90.2          & 69.6          & 66.5          & \textbf{90.0}          & 67.6          & 63.5          & 74.1          & 54.8          & 42.9           \\
\mfvi{}$^{\star}$                                    &                                                                               & 90.2          & 69.7          & 67.1          & 89.7          & 67.4          & 63.4          & 73.3          & 54.7          & 42.5           \\
\hgnn{}$^{\star}$ (our model)          &                                                                               & 90.2          & \textbf{70.7} & \textbf{67.5} & 89.9          & \textbf{68.2} & \textbf{64.2} & \textbf{74.9} & \textbf{55.7} & \textbf{43.6}  \\ 
\midrule
\cite{liu-etal-2022-autoregressive}  & T5$_{3B}$                                                                          & 91.3          & 72.7          & 70.5          & -             & -             & -             & -             & -             & -              \\ 
\midrule
\cite{wang-lu-2020-two}                          & \multirow{9}{*}{\alb{}}                                                       & 89.5          & 67.6          & 64.3          & 88.6          & 63.3          & 59.6          & -             & -             & -              \\
\cite{wang-etal-2021-unire}$^{\star}$                    &                                                                               & 90.2          & -             & 66.0          & 89.5          & -             & 63.0          & -             & -             & -              \\
\cite{zhong-chen-2021-frustratingly}$^{\star}$           &                                                                               & 90.9          & 69.4          & 67.0          & 90.3          & 66.1          & 62.2          & -             & -             & -              \\
\cite{yan-etal-2021-partition}                   &                                                                               & 89.0          & -             & 66.8          & 89.3          & -             & 62.5          & -             & -             & -              \\
\cite{ye-etal-2022-packed}$_{\textit{re}}^{\star}$   &                                                                               & 91.3          & 72.5          & 70.5          & 90.5          & 69.3          & 66.1          & -             & -             & -              \\
\bkb{}$^{\star}$                                &                                                                               & 91.5          & 72.9          & 70.2          & 91.6          & 70.2          & 66.6          & -             & -             & -              \\
\gcn{}$^{\star}$                                     &                                                                               & 91.7          & 73.1          & 69.9          & \textbf{92.0}          & 71.5          & 67.9          & -             & -             & -              \\
\mfvi{}$^{\star}$                                    &                                                                               & 91.6          & 72.7          & 70.1          & 89.9          & 68.5          & 65.1          & -             & -             & -              \\
\hgnn{}$^{\star}$ (our model)                                    &                                                                               & \textbf{91.9} & \textbf{73.5} & \textbf{70.8} & 91.9 & \textbf{71.9} & \textbf{68.3} & -             & -             & -              \\
\bottomrule
\end{tabular}
    \caption{F1 scores and standard deviations on \acef{}, \ace{} and \scierc{}. The models marked with $\star$ leverage cross-sentence information. A model with subscript $\textit{re}$ means we re-evaluate the model with the  evaluation method commonly used in other work\protect\footnotemark. \bkb{}, \mfvi{} and \gcn{} are our baseline models.}
    \label{tab:main}
    % \lessdis{}
\end{table*}
\footnotetext{\citet{ye-etal-2022-packed} count a symmetric relation twice for evaluation which is inconsistent with previous work.}

\paragraph{Datasets}
We experiment on \scierc{} \cite{luan-etal-2018-multi}, \acef{} \cite{doddington-etal-2004-automatic} and \ace{} \cite{ace2005}. We follow \citet{ye-etal-2022-packed} to split \acef{} into 5 folds and split \ace{} and \scierc{} into train/dev/test sets. See Appendix \ref{apx.data} for detailed dataset statistics.

\paragraph{Evaluation metrics}\label{eval}
We report micro labeled F1 measures for NER and RE. For RE, the difference between Rel and Rel+ is that the former requires correct prediction of subject and object entity spans and the relation type between them, while the latter additionally requires correct prediction of subject and object entity types.

\paragraph{Baseline}
Our baseline models include: i) \textbf{Backbone}. It is described in Sect.~\ref{sec:backbone} and does not contain the higher-order interaction module. ii) \textbf{GCN}. It has a similar architecture to \citet{sun-etal-2019-joint,nguyen-etal-2021-cross} and does not contain higher-order hyperedges. See Appendix~\ref{app.gcn} for a detailed description. iii) \textbf{MFVI}. It defines a CRF on the same hypergraph as our model and uses MFVI instead of hypergraph neural networks for higher-order inference. See Appendix~\ref{app.mfvi} for a detailed description. 

\paragraph{Implementation details}
For a fair comparison with previous work, we use \bertbase{}\cite{devlin-etal-2019-bert} and \albert{}\cite{DBLP:conf/iclr/LanCGGSS20} as the base encoders for \acef{} and \ace{}, \scibert{} \cite{beltagy-etal-2019-scibert} as the base encoder for \scierc{}. GCN and MFVI are also built upon Backbone. The implementation details of experiments are in Appendix \ref{apx.imple}.

%--------------------------
\paragraph{Main results}
For \hgnn{}, we report the best results among the following variants of hypergraphs with different types of hyperedges: $\ter{}$, $\cop{}$, $\sib{}$, $\gp{}$, $\tersib{}$, $\tercop{}$, $\tergp{}$, $\tersibcop{}$, $\tersibgp{}$,  $\tercopgp{}$, and $\tersibcopgp{}$. The best variants of \hgnn{}  are \tersibcop{} on \scierc{} and \ace{} (\bertb{}); \tersib{} on \ace{} (\alb{}); \tercop{} on \acef{}. For \mfvi{} we use the same variants as used in \hgnn{}.

Table~\ref{tab:main} shows the main results. Surprisingly, \bkb{} outperforms prior approaches in almost all metrics by a large margin (except on \acef{} with \bertb{} and \ace{} with \alb{}), which we attribute to the reduction of error propagation with a span pruning mechanism. Our proposed model \hgnn{} outperforms almost all baselines in all metrics (except the entity metric on \acef{}), validating that using hyperedges to encode higher-order interactions is effective (compared with \gcn{}) and that using hypergraph neural networks for higher-order modeling and inference is better than CRF-based probabilistic modeling with MFVI.  Finally, we remark that \hgnn{} obtains state-of-the-art performances on all the three datasets.

%============================================================================
\section{Analysis}

\subsection{Effectiveness of the span pruner}\label{sec.eff_pruner}

\begin{table}[!htb]
    % \lessdis{}
    \setlength\tabcolsep{5pt}
    \small
    \centering
    \begin{tabular}{llrrrrrr} 
    \toprule
           &       & \multicolumn{3}{c}{\scierc{}}                                             & \multicolumn{3}{c}{\ace{} (\bertb{})}                                             \\ 
    \cmidrule{3-8}
           &       & \multicolumn{1}{c}{P} & \multicolumn{1}{c}{R} & \multicolumn{1}{c}{F1} & \multicolumn{1}{c}{P} & \multicolumn{1}{c}{R} & \multicolumn{1}{c}{F1}  \\ 
    \midrule
    \multirow{3}{*}{Eid}    & train & 98.8                  & 98.2                  & 98.5                   & 100.0                   & 99.9                  & 99.9                    \\
           & dev   & 81.0                  & 81.6                  & 81.3                   & 94.7                  & 94.6                  & 94.7                    \\
           & test  & 80.4                  & 78.7                  & 79.5                   & 95.6                  & 95.8                  & 95.7                    \\ 
    \midrule
    \multirow{3}{*}{Pruner} & train & 38.0                  & 99.2                  & 54.9                   & 37.2                  & 99.9                  & 54.2                    \\
           & dev   & 38.1                  & 99.1                  & 55.0                   & 36.4                  & 99.7                  & 53.3                    \\
           & test  & 38.7                  & 99.2                  & 55.7                   & 37.0                  & 99.8                  & 54.0                    \\
    \bottomrule
    \end{tabular}
    \caption{Evaluation results on entity existence of the span pruner vs. an entity identifier.}
    \label{tab.ent_pr}
\end{table}

\begin{table}[!htb]
% \lessdis{}
% \small
\centering
    \begin{tabular}{llccc} 
        \toprule
        \multicolumn{2}{l}{\scierc{}}               & Ent  & Rel  & Rel+  \\ 
        \midrule
        \multirow{2}{*}{Eid}    & \bkb{}  & 69.4 & 50.3 & 39.0  \\
                                & \hgnn{} & 69.4 & 51.5 & 39.5  \\ 
        \midrule
        \multirow{2}{*}{pruner} & \bkb{}  & 71.3 & 52.3 & 40.2  \\
                                & \hgnn{} & 74.9 & 55.7 & 43.6  \\ 
        \toprule
        \multicolumn{2}{l}{\ace{} (\bertb{})}       & Ent  & Rel  & Rel+  \\ 
        \midrule
        \multirow{2}{*}{Eid}    & \bkb{}  & 89.5 & 68.3 & 65.3  \\
                                & \hgnn{} & 89.5 & 68.5 & 66.0  \\ 
        \midrule
        \multirow{2}{*}{pruner} & \bkb{}  & 90.0 & 69.8 & 66.7  \\
                                & \hgnn{} & 90.2 & 70.7 & 67.5  \\
        \bottomrule
    \end{tabular}
    \caption{F1 scores of \bkb{} and \hgnn{} with and without a pre-trained span pruner on the \scierc{} and \ace{} (\bertb{}) test set.}
    \label{tab.eid_vs_pruner}
    % \lessdis{}
\end{table}

% First, we report the results of the span pruner on \scierc{} and \ace{} with \bertb{} in Table \ref{tab.ent_pr} for the following analysis. We use the original NER module in PL-marker as an entity identifier (i.e., Eid in Table \ref{tab.ent_pr}) for comparison, which is trained only on entity existence.

% We report the evaluation result of entity existence of span pruner vs. an entity identifier in Table \ref{tab.ent_pr}. We use the NER module in PL-marker as the entity identifier and train the NER module with only the existence of the entities. 

To study the effectiveness of the span pruner, we replace it with an entity identifier which is the original NER module from PL-marker and is trained only on entity existence.
% (i.e., w/o pruner) 
% and experiment on \scierc{} and \ace{} with \bertb{}. 
The performance of the span pruner and the entity identifier (denoted by Eid) on entity existence is shown in Table \ref{tab.ent_pr}. We can observe that if we replace the span pruner with the entity identifier, the recall of gold \emph{unlabeled} entity spans drops from 99.2 to 78.7 on the \scierc{} test set, and drops from 99.8 to 95.8 on the \ace{} test set. 
We further investigate how the choice of the span pruner vs. the entity identifier influences NER and RE performances. The results are shown in Table \ref{tab.eid_vs_pruner}. We can see that without a span pruner, both NER and RE performances drop significantly, validating the usefulness of using a span pruner. Moreover, it has a consequent influence on the higher-order inference module (i.e., HGNN). Without a span pruner, the improvement from using a HGNN over Backbone is marginal compared to that with a span pruner. We posit that without a pruner many gold entity spans could not exist in the hypergraph of HGNNs, making true entities and relations less connected in the hypergraph and thus diminishing the usefulness of HGNNs.

%---------------------------------------------

\subsection{Effect of the choices of hyperedges}\label{sec.edge}

\begin{table}[!htb]
    % \lessdis{}
    % \small
    \centering
    \begin{tabular}{llll}
        \toprule
        \multirow{2}{*}{\hgnn{}}                         & \multicolumn{3}{c}{\scierc{}}                                \\ 
        \cline{2-4}
                                        & \multicolumn{1}{c}{\ent{}} & \multicolumn{1}{c}{\rel{}} & \multicolumn{1}{c}{\relp{}}  \\ 
        \midrule
        \bkb{}         & 71.3                       &52.3                & 40.2                      \\
        \ter{}         & 74.2                       &55.1                & 42.6                      \\
        \sib{}         & 71.7                       &54.3                & 41.7                      \\
        \cop{}         & 71.7                       &52.9                & 40.8                      \\
        \gp{}          & 71.3                       &51.9                & 40.1                      \\
        \tersib{}      & 74.7                       &\textbf{55.9}       & 43.3                     \\
        \tercop{}      & 74.7                       &55.7                & \textbf{43.6}             \\
        \tergp{}       & 74.5                       &54.9                & 42.4                      \\
        \tersibcop{}   & \textbf{74.9}              &55.7                & \textbf{43.6}                   \\
        \tersibgp{}    & 74.2                       &54.1                & 41.8                      \\
        \tercopgp{}    & 74.7                       &54.0                & 42.3                      \\
        \tersibcopgp{} & 74.3                       &54.6                & 41.7                      \\
        \bottomrule
    \end{tabular}
    \caption{F1 scores of \hgnn{} with different graph topologies on the \scierc{} test set.}
    \label{tab:ana_topo}
    % \lessdis{}
\end{table}

We compare different variants of HGNN with different combinations of hyperedges.  Note that if $\ter{}$ is not used, entity nodes do not have any hyperedges connecting to them, so their representations would not be refined. We can see that in the $\sib{}$ and $\cop{}$ variants, the NER performance improves slightly, which we attribute to the shared encoder of NER and RE tasks~\footnote{Though ~\citet{zhong-chen-2021-frustratingly} argue that using shared encoders would suffer from the \emph{feature confusion problem}, later works show that shared encoders can still outperform separated encoders~\cite{yan-etal-2021-partition,yan-etal-2022-empirical}.}. On the other hand, in the $\ter{}$ variant, entity node representations are iteratively refined, resulting in significantly better NER performance than $\bkb{}$ (74.2 vs. 71.3). Combining $\ter{}$ edges with other rel-rel edges (e.g., $\sib{}$) is generally better than using $\ter{}$ alone in terms of NER performance, suggesting that joint (and higher-order) modeling of NER and RE indeed has a positive influence on NER, while prior pipeline approaches (e.g., PL-marker) cannot enjoy the benefit of such joint modeling.  

For RE, $\sib{}$ and $\cop{}$ have positive effects on the performance (despite $\gp{}$ having a negative effect somehow), showing the advantage of modeling interactions between two different relations. Further combining them with $\ter{}$ improves RE performances in all cases, indicating that NER also has a positive effect on RE and confirming again the advantage of joint modeling of NER and RE.

%-------------------------------------------------------

\subsection{Inference speed of higher-order module}
To analyze the computing cost of our higher-order module, we present the inference speed of \hgnn{} with three baseline models \bkb{},  \gcn{} and \mfvi{} on the test sets of \scierc{} and \ace{}. Inference speed is measured by the number of candidate entities processed per second. The results are shown in Table \ref{tab.speed}. We can observe that when utilizing a relatively smaller PLM, \hgnn{}, \gcn{} and \mfvi{} were slightly slower than the first-order model \bkb{}. However, the difference in speed between \hgnn{} and the other models was relatively small. When using \alb{}, which is much slower than \bertb{}, all four models demonstrated comparable inference speeds.

\begin{table}[!htb]
% \small
\centering
\begin{tabular}{lccc} 
\toprule
         & \multicolumn{1}{c}{\scierc{}}  & \multicolumn{2}{c}{\ace{}}                                   \\ 
\cline{2-4}
         & \multicolumn{1}{c}{\scib{}} & \multicolumn{1}{c}{\bertb{}} & \multicolumn{1}{c}{\alb{}}  \\ 
\midrule
\bkb{} & 19.4                        & 38.0                          & 6.1                         \\
\gcn{}      & 15.7                        & 33.8                          & 6.3                         \\
\mfvi{}     & 16.5                        & 36.9                          & 6.1                         \\
\hgnn{}     & 15.7                        & 30.7                          & 6.0                         \\
\bottomrule
\end{tabular}
\caption{Comparison of inference speed (\#entities/sec) between \hgnn{} and three baseline models on test sets of \scierc{} and \ace{}.}
\label{tab.speed}
\end{table}

%-------------------------------------------

\subsection{Error correction analysis}

\begin{figure}[!ht]
\lessdis{}
\centering
    \subfigure[Error correction matrix of \hgnn{} vs. \bkb{} of entities]
        {                    
        \begin{minipage}[t]{0.95\columnwidth} 
        \centering                                           
        \includegraphics[width=\linewidth]{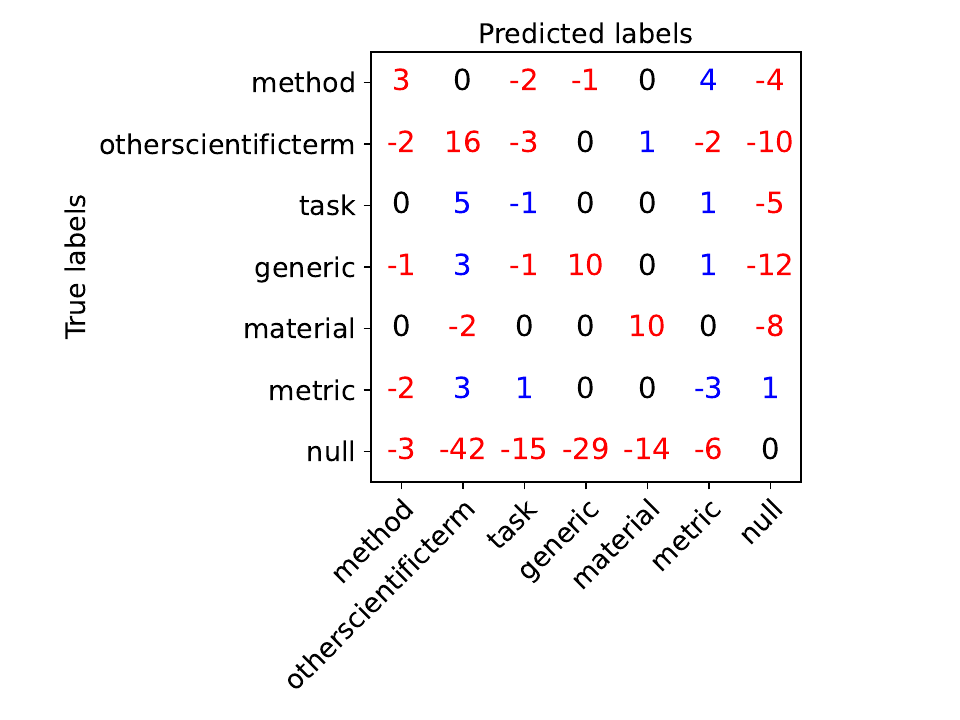}
        \end{minipage}
        }
    \subfigure[Error correction matrix of \hgnn{} vs. \bkb{} of relations]
        {                    
        \begin{minipage}[t]{0.95\columnwidth} 
        \centering                                       
        \includegraphics[width=\linewidth]{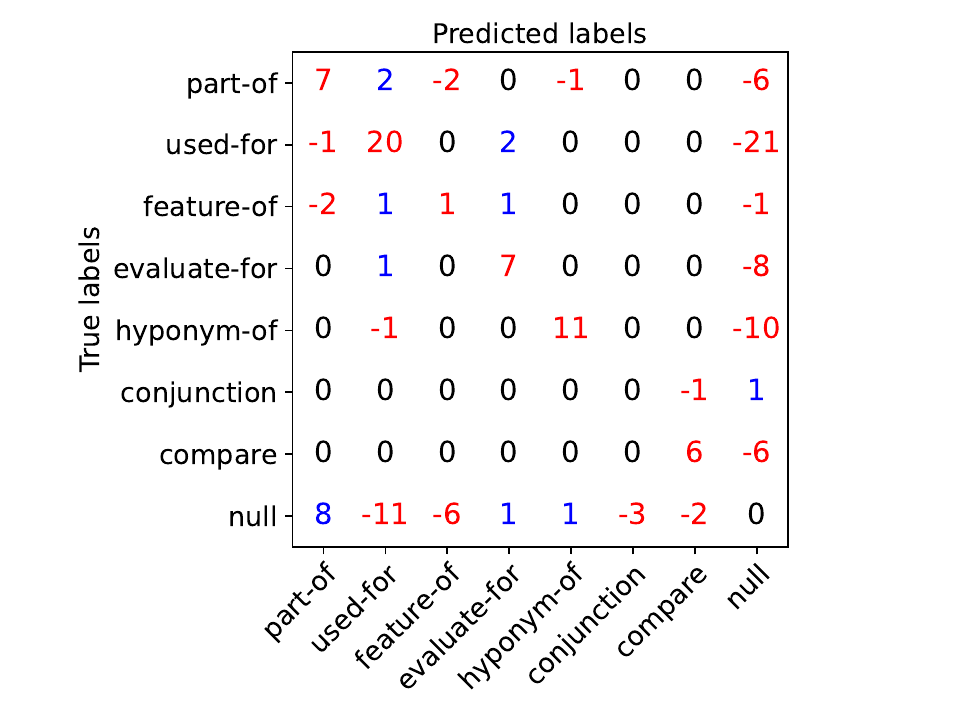}
        \end{minipage}
        }
\caption{Error correction of entity and relation types on the \scierc{} dataset. Red color indicates positive corrections and blue color indicates negative corrections. 
Specifically, positive numbers on the diagonal of the matrix (in red color) indicate that \hgnn{} makes more correct predictions compare to \bkb{}; negative numbers on non-diagonal entries (in red color) indicate that \hgnn{} makes fewer wrong predictions compare to \bkb{}. Numbers in blue indicate the opposite. We do not count the null-null case.}
\label{fig.error_correct}
% \lessdis{}
\end{figure}

We provide quantitative error correction analysis between our higher-order approach \hgnn{} and the first-order baseline \bkb{} on the SciERC dataset in Fig. \ref{fig.error_correct}. We can see that most error corrections of entities and relations made by \hgnn{} come from two categories. The first category is where \bkb{} incorrectly predicts a true entity or relation as null, and the second category is where \bkb{} incorrectly assigns a label to a null sample.

%========================================================================
\section{Related Work}

\paragraph{Entity and relation extraction}
The entity and relation extraction task has been studied for a long time. The mainstream methods could be divided into pipeline and joint approaches. Pipeline methods tackle the two subtasks, named entity recognition and relation extraction, consecutively \cite{zelenko2003kernel, chan-roth-2011-exploiting, zhong-chen-2021-frustratingly,ye-etal-2022-packed}. By utilizing a new marker-based embedding method, \citet{ye-etal-2022-packed} becomes the new state-of-the-art ERE model. However, pipeline models have the inherent error propagation problem and they could not fully leverage interactions across the two subtasks. Joint approaches, on the other hand, can alleviate the problem by simultaneously tackling the two subtasks, as empirically revealed by \citet{yan-etal-2022-empirical}. Various joint approaches have been proposed to tackle ERE.  \citet{miwa-bansal-2016-end, katiyar-cardie-2017-going} use a stacked model for joint learning through shared parameters. \citet{miwa-sasaki-2014-modeling, gupta-etal-2016-table, wang-lu-2020-two, wang-etal-2021-unire, yan-etal-2021-partition} tackle both the NER and RE tasks as tagging entries of a table. \citet{fu-etal-2019-graphrel,sun-etal-2019-joint} leverage a graph convolutional network (GCN) on an instance dependency graph to enhance instance representations. \cite{nguyen-etal-2021-cross} propose a framework to tackle multiple Information Extraction tasks jointly including the ERE task where a GCN is used to capture the interactions between related instances. 

Another line of research is based on text-to-text models for structure prediction including ERE. Normally they are not task-specialized and could solve several structure prediction tasks in a unified way  \cite{paolini2021structured,lu-etal-2022-unified,liu-etal-2022-autoregressive}.

This work is similar to \citet{sun-etal-2019-joint,nguyen-etal-2021-cross} for we both use a graph neural network to enhance the instance representations. The main difference is that the GCN they use cannot adequately model higher-order relationship among multiple instances, while our hypergraph neural network is designed for higher-order modeling.

\paragraph{CRF-based higher-order model}
A commonly used higher-order model utilizes approximate inference algorithms (mean-field variational inference or loopy belief propagation) on CRFs. \citet{zheng2015conditional} formulate the mean-field variational inference algorithm on CRFs as a stack of recurrent neural network layers, leading to an end-to-end model for training and inference. Many higher-order models employ this technique for various NLP tasks, such as semantic parsing \cite{wang-etal-2019-second,wang-tu-2020-second} and information extraction \cite{Jia2022JointIE}.

\paragraph{Hypergraph neural network}
Hypergraph neural network (HyperGNN) is another way to construct an higher-order model. Traditional Graph Neural Networks employ pairwise connections among nodes, whereas HyperGNNs use a hypergraph structure for data modeling. \citet{feng2019hypergraph} and \citet{bai2021hypergraph} proposed spectral-based HyperGNNs utilizing the normalized hypergraph Laplacian. \citet{arya2020hypersage} is a spatial-based HyperGNN which aggregates messages in a two-stage procedure. \citet{ijcai2021p0353} proposed \textit{UniGNN}, a unified framework for interpreting the message passing process in HyperGNN. \citet{9795251} introduced a general high-order multi-modal data correlation modeling framework to learn an optimal representation in a single hypergraph based framework. 
% Both frameworks include a two-stage message passing process.

%===============================================================================

\section{Conclusion}
In this paper, we present \hgnn{}, a joint entity and relation extraction model equipped with a span pruning mechanism and a higher-order interaction module (i.e., HGNN).
% (i.e., hypergraph neural network (HGNN)).
We found that simply using the span pruning mechanism by itself greatly improve the performance over prior state-of-the-art PL-marker, indicating the existence of the error propagation problem for pipeline methods. We compared our model with prior tranditional GNN-based models which do not contain hyperedges connecting multiple instances and showed the improvement, suggesting that modeling higher-order interactions between multiple instances is beneficial. Finally, we compared our model with the most popular higher-order CRF models with MFVI and showed the advantages of HGNN in higher-order modeling.

%===============================================================================

\section*{Limitations}
Our model achieves a significant improvement in most cases (on \acef{}, \scierc{} datasets and on \ace{} with Bert$_{\text{base}}$).
While on \ace{} with stronger encoder (e.g., ALBERT) we observe less siginificant improvements. We posit that, with powerful encoders, the recall of gold entity spans would increase, thereby mitigating the error propagation issue and diminishing the benefit of using a span pruning mechanism. 

Another concern regarding our model is computational efficiency. The time complexity of the \emph{Subject-oriented Packing for Span Pair} encoding scheme from  PL-marker grows linearly with the size of candidate span size.  Recall that we over-predict many spans using a span pruning mechanism, which slows down the running time. In practice, our model's running time is around as three times as that of PL-marker.

\section*{Acknowledgments}
% We thank the anonymous reviewers for their constructive comments. 
This work was supported by the National Natural Science Foundation of China (61976139).

\bibliography{anthology,custom}
\bibliographystyle{acl_natbib}

\appendix

\section{Appendix}
\label{sec:appendix}

%---------------------------------------------
\subsection{Datasets}\label{apx.data}
We use \acef{}, \ace{} and \scierc{} datasets in our experiments, the data statistics of each dataset is shown in Table \ref{tab.data}.

\begin{table}[!ht]
    \centering
    \begin{tabular}{lccc} 
        \toprule
                & \#sent & \#entity & \#relation  \\ 
        \midrule
        \acef{} & 8683   & 22735    & 4087        \\
        \ace{} & 14525  & 38287    & 7070        \\
        \scierc{}  & 2687   & 8094     & 4648        \\
        \bottomrule
        \end{tabular}
    \caption{The statistic of datasets}
    \label{tab.data}
\end{table}

\subsection{Bidirectional prediction of RE}
Following previous work \cite{DBLP:conf/ecai/EbertsU20, ye-etal-2022-packed}, we establish an inverse relation for each asymmetric relation for a bidirectional prediction. The model can learn the inverse relations of asymmetric relations and improve the performance in this way.

\subsection{Implementation details}\label{apx.imple}
 % We use a PLM encoder of \scibert{} \cite{beltagy-etal-2019-scibert} for \scierc{}, \bert{} \cite{devlin-etal-2019-bert} and \albert{} \cite{DBLP:conf/iclr/LanCGGSS20} for \ace{}.
 We adopt the same cross-sentence information incorporating method used in \citep{zhong-chen-2021-frustratingly, ye-etal-2022-packed} which extend the original sentence to a fixed window size $W$ with its left and right context. We set $W=512$ for \scierc{}, $W=384$ for \acef{} and $W=256$ for \ace{}. For the pruner training and inference, we consider the span length limitation $L$ of 12 for \scierc{} and 8 for \acef{} and \ace{}. For pruners of any datasets and PLMs, the top-K ratio $\lambda=0.5$, the boundaries of $K$ are $l_\min =3, l_\max=18$.  We use three hypergraph convolution layers for \gcn{}, \mfvi{} and \hgnn{}. As the entity recall is high enough, 
 pruners use on \acef{} and \ace{} are only trained with \bertb{}. For all experiments, we run each configuration with 5 different seeds and report the average micro-F1 scores and standard deviation. 
 
 For the pruner, the output sizes of $\FFN_{\ST{}}, \FFN_{\ED{}}$ and $\FFN_q$ are  $d_m=768$, the bi-affine embedding size $d_\biaf{}=256$, the output size of $\FFN_{\attn}$ is 256.
 
For the backbone module, the output sizes of $\FFN_s, \FFN_o$ and $\FFN_r$ are tuned on $[400, 512, 768]$ for 
all datasets.

For the hypergraph neural network, the output sizes of $\FFN^{ter}_r, \FFN^{ter}_s, \FFN^{ter}_o, \FFN^{\ztt}_a, \FFN^{\ztt}_b$ are tuned among $[256, 400, 512]$ and fixed on 400 for all experiments on \scierc{}. The output sizes of $\FFN^{ter}_e, \FFN^{\ztt}_e$ are tune on $[256, 400, 512, 768]$ for all experiments.
For \gcn{}, \mfvi{} and \hgnn{}, we all use three layers to refine the node representations.
We train our models with Adam optimizer and a liner scheduler with warmup ratio of 0.1. We tune the eps of Adam optimizer on $[1e-8, 1e-9]$ for \ace{}, and eps=$1e-8$ for other datasets.
The batch size of all experiments are 18. The learning rate of PLM are $2e-5$, for other module the learning rate is tune on $[5e-5, 1e-4]$.
The epochs on \scierc{} for \bkb{}  are 20, and 30 for other models. The epochs on \acef{} and \ace{} (\bertb{}) are 15, on \acef{} and \ace{} (\alb{}) are 10.
We do all experiments on a A40 GPU with apex fp16 training option on.

%----------------------------------------------------------------------

\subsection{Details of the span pruner}
\label{appd:pruner}
We obtain contextualized representations of the tokens $\rvx$ and levitated marker representations $\rvxs$ (for $[O]$) and $\rvxe$ (for $[\bso]$) . Then we concatenate two kinds of span representations---bi-affine \cite{Dozat2016DeepBA} and attentive pooling---as the final one. For a span $s_i$ consisting of tokens $x_{\ST(i)},...,x_{\ED(i)}$, its bi-affine span representation is a $d_\biaf$-dimension vector,
\begin{align*}
    & \rvh_{\ST}(s_i)=\FFN_{\ST}(\rvx_{\ST(i)};\rvxs(i))\\
    & \rvh_{\ED}(s_i)=\FFN_{\ED}(\rvx_{\ED(i)};\rvxe(i))\\
    & \rvh_\biaf(s_i)=[\rvh_{\ST}(s_i);1]^\top \emW_p [\rvh_{\ED}(s_i);1]
\end{align*}
the symbol $;$ is the concatenation operation, $\FFN_{\ST}$ and $\FFN_{\ED}$ are feed-forward layers with an output size $d_m$ and $\emW_p\in \mathbb{R}^{(d_m+1)*d_{\biaf}*(d_m+1)}$ is a learn-able weight. The attentive pooling layer is a weighted average over the contextualize token representations in the span,
\begin{align*}
 & w_j = \FFN_{q}(\rvx_j); w_j = \frac{\exp{w_j}}{\sum_{\ST(i)\le l\le \ED(i)}\exp{w_l}} \\
 & \rvh_\attn(s_i)=\sum_{\ST(i)\le j\le \ED(i)}w_j \rvx_j
\end{align*}
and the final span representation is,
$$
    \rvh(s_i)=\FFN_{\attn}[\rvh_\biaf(s_i);\rvh_\attn(s_i)]
$$

\paragraph{Training and Inference}
Given the gold binary tag $y(s_i) \in \{0, 1\}$ (indicating the existence of a candidate span in the gold span set), we train the span pruner with the binary cross-entropy (BCE) loss:
\[
    \hat{y}(s_i)=\sig(\FFN(\rvh(s_i)))
\]
\begin{align*}
    L = -\sum_{1\leq i \leq m}  & [y(s_i)\log\hat{y}(s_i) \\
                                & + (1-y(s_i))(1-\log\hat{y}(s_i) ]
\end{align*}

\subsection{Mean-field Variant Inference}\label{app.mfvi}
Here we introduce the method used in baseline \mfvi{}. The hyperedges in our graph are replaced by factors in \mfvi{}, so there are also four kinds of factors: \ter{}, \sib{}, \cop{}, \gp{}. 

\paragraph{first-order scores}
We use the node representations to score the entities and relations for each label (include the \nul{}).
\begin{align*}
    \rvu^s_i & =\FFN^u_s(\rvg(v_s^i)) \\
    \rvu^o_j & =\FFN^u_o(\rvg(v_o^j)) \\
    \rvu^r_{ij} & =\FFN^u_r(\rvg(v_r^{ij})) \\
\end{align*}
$\rvu^s, \rvu^o\in R^{|\nertype{}|+1}$, $\rvu^r\in R^{|\retype{}|+1}$.

\paragraph{Higher-order scores}
Each factor scores the joint distribution of the node types connected to it. For a \ter{} factor connects a subject $v_s^i$, an object $v_o^j$ and a relation $v_r^{ij}$, the factor score $\rvf^{ter}_{ij}\in R^{|\nertype{}+1|^2 |\retype{}+1|} $ is:
\begin{align*}
    \rvh^s_i &= \FFN^{ter}_s(\rvg(v_s^i)) \\
    \rvh^o_j &= \FFN^{ter}_o(\rvg(v_o^j)) \\
    \rvh^r_{ij} &= \FFN^{ter}_r(\rvg(v_r^{ij})) \\
    \rvf^{ter}_{ij} & = \FFN^{ter}_f(\rvh^s_i \circ \rvh^o_j \circ \rvh^r_{ij})
\end{align*}

For a factor $\ztt{}, z\in \{\sib{}, \cop{}, \gp{}\}$which connects two relations, we name them relation $a$ and $b$ for simplicity. If relation $a$ is $v_r^{ij}$, then relation $b$ is $v_r^{ik}, v_r^{kj}$ and $v_r^{jk}$ for \sib{}, \cop{} and \gp{} respectively. We use $\rvg(a), \rvg(b)$ to refer to the relation representations of relations $a$ and $b$.
The factor score $\rvf^{\ztt{}}_{ijk}\in R^{|\retype{}+1|^2}$ is defined as:
\begin{align*}
    \rvh_a &= \FFN^{ter}_s(\rvg(a)) \\
    \rvh_b &= \FFN^{ter}_o(\rvg(b)) \\
    \rvf^{\ztt}_{ijk} & = \FFN^{ter}_f(\rvh_a \circ \rvh_b)
\end{align*}

\paragraph{higher-order inference}
In the model, computing the node distribution can be seen as doing posterior inference on a Conditional Random Field (CRF). MFVI iteratively updates a factorized variational distribution Q to approximate the posterior label distribution. We use $Q_{s_i}(e_1), Q_{o_j}(e_2)$ to refer to the probability of  subject $v_s^i$ and object $v_o^j$ has entity type $e_1$ and $e_2$ respectively and  $Q_{r_{ij}}(r)$ represents the relation $v_r^{ij}$ has the relation type $r$. For simplicity, we use $u^s_i(e_1), u^o_j(e_2), u^r_{ij}(r_1), f^{ter}_{ij}(e_1, e_2, r_1), f^{\ztt}_{ijk}(r_1, r_2) $ to represent the first-order and higher-order scores when the subject $v_s^i$, the object $v_o^j$ have entity type $e_1, e_2$, the relation $a$ ($v_r^{ij}$), the relation $b$ have relation types $r_1, r_2$ respectively.
Following is the iterately updating of the distribution $Q_s, Q_o, Q_r$.
For a subject $v_s^i$. The message only passed from \ter{} factor in the $l$-th iteration is:
\begin{align*}
    & F_{s_i}^l(e_1)= \\
    & \sum_j\sum_{e_2}Q_{o_j}^{l-1}(e_2)(\sum_{l_1}Q_{r_{ij}}^{l-1}(r_1)f^{ter}_{ij}(e_1, e_2, r_1))
\end{align*}
similarly, the message passed from \ter{} factor to the object $v_o^j$ is:
\begin{align*}
    & F_{o_j}^l(e_2)=\\
    & \sum_i\sum_{e_1}Q_{s_i}^{l-1}(e_1)(\sum_{l_1}Q_{r_{ij}}^{l-1}(r_1)f^{ter}_{ij}(e_1, e_2, r_1))
\end{align*}
For a relation $v_r^{ij}$, the message could be passed from four factors, we list them by the source. 
From \ter{} factor:
\begin{align*}
    & T_{r_{ij}}^l(r_1) = \\
    & \sum_{e_1}\sum_{e_2} Q_{s_i}^{l-1}(e_1)Q_{o_j}^{l-1}(e_2)f_{ij}^{ter}(e_1, e_2,r_1)
\end{align*}
From \sib{} factor:
\begin{align*}
    & S_{r_{ij}}^l(r_1) = \\ 
    & \sum_k \sum_{r_2} Q_{r_{ik}}^{l-1}(r_2)(f_{ijk}^{sib}(r_1, r_2)+f_{ikj}^{sib}(r_2, r_1))
\end{align*}
From the \cop{} factor:
\begin{align*}
    & C_{r_{ij}}^l(r_1) = \\ &\sum_k\sum_{r_2} Q_{r_{kj}}^{l-1}(r_2)(f_{ijk}^{cop}(r_1, r_2)+f_{kji}^{cop}(r_2, r_1))
\end{align*}
From the \gp{} factor:
\begin{align*}
    G_{r_{ij}}^l(r_1) & = \\ & \sum_k\sum_{r_2} (Q_{r_{jk}}^{l-1}(r_2)f_{ijk}^{gp}(r_1, r_2) \\ &+Q_{r_{ki}}^{l-1}(r_2)f_{kij}^{gp}(r_2, r_1))
\end{align*}
The posterior distribution of entity $e_i$ with respect to the subject $s_i$ and object $o_i$ :
\begin{align*}
    Q_{s_i}^l(e) \propto \exp(u^s_i(e) + F_{s_i}^l(e))\\
    Q_{o_j}^l(e) \propto \exp(u^o_j(e) + F_{o_j}^l(e))
\end{align*}
Then the entity distribution is : $Q^l_{s_i}+Q^l_{o_i}$

We initial the Q of subject $v^s_i$, object $v^o_j$, the relation $v^r_{ij}$ by normalizing the unary potential $\exp(u^s_i), \exp(u^o_j), \exp(u^r_{ij})$ respectively.
The posterior distribution of the relation $r_{ij}$ is:
\begin{align*}
    Q_{r_{ij}}^l(r) & \propto \exp( u^r_{ij}(r) + \mathds{1}_{ter}T_{r_{ij}}^l(r) \\ & \mathds{1}_{sib}S_{r_{ij}}^l(r) + \mathds{1}_{cop}C_{r_{ij}}^l(r) + \mathds{1}_{gp}G_{r_{ij}}^l(r))
\end{align*}
The symbol $\mathds{1}_{\ztt{}}$, $z\in \{\ter{}, \sib{}, \cop{}, \gp{}\}$ indicates whether the factor $\ztt{}$ exists in the graph.

%-------------------------------------------------
\subsection{GCN}\label{app.gcn}
Here is the introduction of the baseline \gcn{}.
As in \hgnn{}, we also build the graph $\gG=(\gV, \gE)$ with subject, object and relation nodes, $\gV=\gV_s\bigcup \gV_o \bigcup \gV_r$. For each relation node $v^{ij}_r\in \gV_r$, we build two edges connecting its subject node $v_s^i\in\gV_s$ and object node $v_o^j\in \gV_o$ respectively. 
\begin{figure}[!ht]
    \centering
    \includegraphics[width=0.45\textwidth]{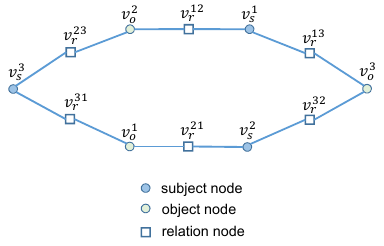} 
    \caption{Illustration of an example graph of \gcn{} }
    \label{fig.recall} 
\end{figure}
the model use $l$ convolution layers to update the node representations. We define the neighbor set $\gN(v)$ of a node $v$ is the nodes connected to it. The node representation update of $l$-th layer is as follow:
\begin{align*}
 & \beta^l(v_1, v) = \rvw^{\top}\sigma(\rvW[\rvg^{l-1}(v_1); \rvm^{l-1}(v)]  \\
   & \alpha^l(v_1, v) = \frac{\exp{\beta^l(v_1, v)}}{\sum_{v_1\in \gN_(v)}\exp{\beta^l(v_1, v)}} \\
   & \rvg^{l}(v) =  \rvg^{l-1}(v) + \sum_{v_1\in \gN(v)} \alpha(v_1, v)\rvg^{l-1}(v_1) \\
\end{align*}

%-----------------------------------------------------
\subsection{Performance with part of the training data}

\begin{table}[th!]
    \centering
    \begin{tabular}{llccc} 
        \toprule
        \multirow{2}{*}{ratio} & \multirow{2}{*}{model} & \multicolumn{3}{c}{ACE2005 (\bertb{})}  \\ 
        \cline{3-5}
                               &                        & Ent  & Rel  & Rel+           \\ 
        \midrule
        \multirow{2}{*}{5\%}   & \bkb{}                    & 80.4 & 39.7 & 36.0           \\
                               & \hgnn{}                   & 79.5 & 42.0 & 38.1           \\ 
        \hline
        \multirow{2}{*}{10\%}  & \bkb{}                    & 83.9 & 51.3 & 47.2           \\
                               & \hgnn{}                   & 84.2 & 53.3 & 49.4           \\ 
        \hline
        \multirow{2}{*}{100\%} & \bkb{}                    & 90.0 & 69.8 & 66.7           \\
                               & \hgnn{}                   & 90.2 & 70.1 & 67.3           \\
        \bottomrule
    \end{tabular}
    \caption{F1 score of \hgnn{} on \ace{} test set when only provide 5\% and 10\% training samples.}
    \label{tab:subset}
\end{table}

From the main results, we can see that the \hgnn{} shows a significantly greater improvement in performance compared to the \bkb{} model on the \scierc{} dataset than on the \ace{} dataset. We guess one of the reason is the size of the training data. Because with more training data, models could learn enough knowledge from a large number of samples and reduce the demand of higher-order information. 
So we compare \hgnn{} to \bkb{} with 5\% and 10\% of training data on the \ace{} (\bertb{}) to see if higher-order inference is more effectiveness with small training data. From the results shown in Table \ref{tab:subset} we can see that the increments of absolute F1 score on \relp{} metric  from \bkb{} to \hgnn{} are 2.1\%, 2.2\% on 5\% and 10\% of training set respectively, which are much higher than 0.6\% on full training set.

%-------------------------------------------------------------------
\subsection{Effect of the number of HGNN layers}
\begin{figure}[!ht]
    \centering
    \includegraphics[width=0.45\textwidth]{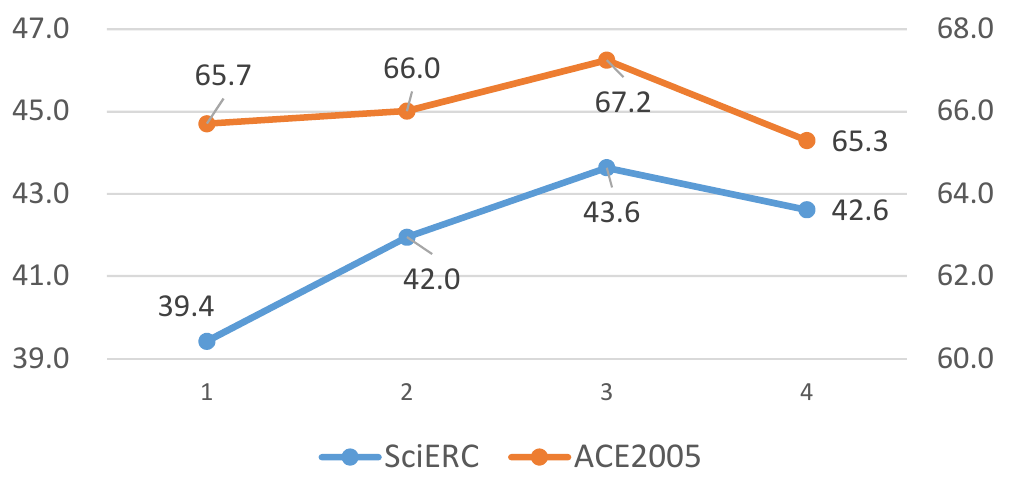} 
    \caption{The change of F1 scores with respect to the number of HGNN layers .}
    \label{fig.iter} 
\end{figure}

From Fig.\ref{fig.iter} we can see that using three HGNN layers performs the best while more layers lead to worse results. We posit that this is because using more HGNN layers would suffer from the well-known over-smoothing problem \cite{Cai2020ANO}. 

%-------------------------------------------------------------------
\subsection{Effect of the aggregation function in message passing}\label{sec.agg}

\begin{table}
    \centering
    \begin{tabular}{lccc} 
        \toprule
             & \multicolumn{3}{c}{\scierc{}}  \\ 
        \cline{2-4}
             & Ent  & Rel  & Rel+          \\ 
        \midrule
        max  & 74.0 & 54.7 & 41.4          \\
        sum  & 73.6 & 54.5 & 41.5          \\
        attn & 74.9 & 55.7 & 43.6          \\
        \bottomrule
    \end{tabular}
    \caption{F1 scores of \hgnn{} (the \tersibcop{} variant) with different aggregation functions on the \scierc{} test set.}
    \label{tab.agg}
\end{table}

% \begin{table}
% \small
% \centering
% \begin{tabular}{c|lll}
% \toprule
%      & \multicolumn{3}{c}{\scierc{}}                                                    \\
% \cline{2-4}
%      & \multicolumn{1}{c}{\ent{}} & \multicolumn{1}{c}{\rel{}} & \multicolumn{1}{c}{\relp{}}  \\
% \hline
% max  & $74.0_{\pm0.4}$         & $54.7_{\pm0.5}$         & $41.4_{\pm0.4}$           \\ 
% sum  & $73.6_{\pm0.6}$         & $54.5_{\pm1.0}$         & $41.5_{\pm0.9}$           \\
% attn & $74.9_{\pm0.5}$         & $55.7_{\pm1.0}$         & $43.6_{\pm0.7}$           \\
% \bottomrule
% \end{tabular}
% \caption{F1 scores of \hgnn{} (the \tersibcop{} variant) with different aggregation functions on the \scierc{} test set.}
% \label{tab.agg}
% \end{table}

We study the influence of using different message aggregation functions. \hgnn{} uses an attention mechanism (\texttt{attn}) to update node representations while it is also possible to use max-pooling (\texttt{max}) or sum-pooling (\texttt{sum}). Table \ref{tab.agg} shows that \texttt{attn} performs the best.

\end{document}